\documentclass[letterpaper, 10 pt, conference]{ieeeconf}
\usepackage[T1]{fontenc}
\IEEEoverridecommandlockouts
\usepackage{cite}
\usepackage{amsmath,amssymb,amsfonts}
\usepackage{algorithmic}
\usepackage{graphicx}
\usepackage{textcomp}
\usepackage{xcolor}
\usepackage{stfloats}   
\usepackage{booktabs}
\def\BibTeX{{\rm B\kern-.05em{\sc i\kern-.025em b}\kern-.08em
    T\kern-.1667em\lower.7ex\hbox{E}\kern-.125emX}}
\begin{document}

\title{A Monolithic Force-Proprioception Soft Acutuator Enabled by Single-Material 3D printing\\
}
\author{
Nan Huang$^{1}$,
Lele Liu$^{2}$,
Junfeng Lu$^{2}$,
Yipan Zhu$^{2}$,
Jiansheng Dai$^{1}$,
Sicong Liu$^{2}$
\\
$^{1}$Department of Mechanical and Energy Engineering,
Southern University of Science and Technology,
Shenzhen, China
\\
$^{2}$Sino-German College of Intelligent Manufacturing,
Shenzhen Technology University,
Shenzhen, China
}
\maketitle
\begin{abstract}

Pneumatic proprioceptive actuators integrate actuation and sensing for soft robots that attract interest due to functional potential. Existing approaches often suffer from assembly errors or stress concentrations caused by heterogeneous materials. In this work, we propose the Monolithic Force-Proprioception Soft (MFPS) design and fabrication method that integrates an Asymmetric Origami Bending (AOB) chamber and a Force-Proprioception Soft (FPS) sensor with single material through one-step Fused Deposition Modeling (FDM) fabrication. Based on the resistance response to strain of conductive thermoplastic polyurethane (TPU), we design and analyze the structure of the FPS sensor, and conduct parametric analysis on the sensing characteristics. The FDM fabrication parameters of the MFPS actuator are analyzed, followed by actuator fabrication and characterization of the actuation and proprioception performance. Experimental results show that the MFPS actuator achieves a bending angle of $40^\circ$, an output force of $12.5$~N, and a resistance change of $26.9\%$ as the applied external force increased from 0 to 45~N. A two-finger force-proprioception gripper is developed based on the MFPS actuator. The grasping and force-proprioception capabilities are experimentally validated, proving that the MFPS design method provides a new approach for the development of self-sensing actuators.

\end{abstract}

\section{Introduction}
Soft robotics has attracted increasing attention due to its compliance, lightweight structure, and potential for safe interaction with humans [1]-[3]. These characteristics make soft robots promising for applications including manipulation, rehabilitation, and human-robot interaction [4]-[6]. Pneumatic soft actuators have therefore been extensively investigated to meet the requirements for these applications. Among them, pneumatic proprioceptive soft actuators have demonstrated promising performance by integrating actuation and sensing capabilities. Especially, soft sensors capable of accommodating the compliance and large deformation of soft actuators have received considerable attention [7]-[9].

For integration, soft sensors can be attached to or embedded within the soft actuator [10], [11]. For instance, soft sensors are attached to soft actuators to monitor their deformation and motion [12], [13]. Studies embedded the soft sensor within the actuators, thereby enhancing the function integration [14]-[16]. Attaching soft sensors onto the actuator surface provides a simple approach to sensing actuator deformation, while embedded sensors enable closer mechanical coupling with the actuator. However, these approaches generally require additional assembly or integration processes, which introduce installation errors and increase fabrication complexity. Therefore, monolithic integration of soft sensors and actuators emerges as a promising approach to simplify fabrication and improve structural compactness.

\begin{figure}[t]
  \centering
  \includegraphics[width=\columnwidth]{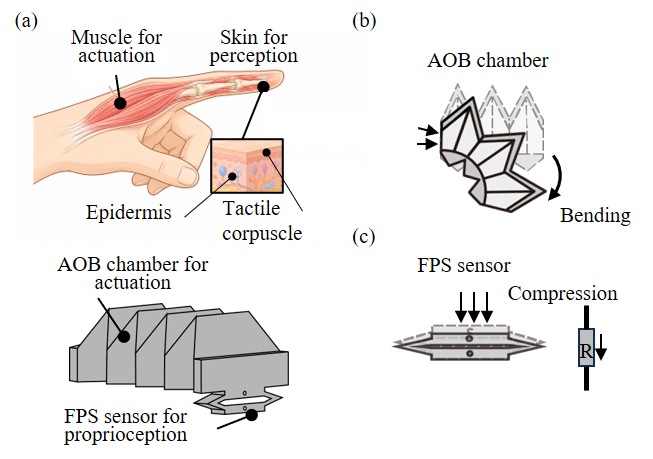}
  \caption{The concept of the MFPS actuator, (a) inspired by a human finger, integrates (b) the pneumatic chamber for bending motion and (c) the force-proprioception sensor within a monolithic structure. }
\end{figure}

Recent advances in 3D-printing fabrication enables the monolithic fabrication of soft actuators and sensors with integrated functionality [17]-[21]. Multimaterial 3D-printing is employed to fabricate soft actuators with embedded strain sensors [17], and sensor-integrated soft grippers with monolithic structures [19]. However, existing approaches often rely on multimaterial printing to combine the distinct properties required for actuation and sensing. The use of multiple materials introduces additional fabrication complexity and potential stress concentration at the material interfaces, which hinders advancements in the mechanical performance and the reliability. Multimaterial printing introduces additional requirements for material compatibility and printer capability. Therefore, a single-material additive fabrication approach that enables both actuation and sensing within a monolithic structure could further simplify fabrication and improve structural integration.

According to transduction mechanisms, soft sensors include capacitive, optical, and resistive types [7]-[9]. Capacitive sensors detect changes in capacitance and offer low power consumption [7]. Optical sensors transduce deformation through changes in light intensity, wavelength, or phase, providing immunity to electromagnetic interference [22],[23]. Resistive sensors convert mechanical deformation into changes in electrical resistance and feature a simple structure [24]. This characteristics make resistive sensors particularly suitable for monolithic integration with soft actuators using a single material, enabling integrated actuation and sensing through a single-step fabrication process.

Single-material, single-step fabrication of soft actuators with deformation proprioception has been previously demonstrated. A PneuNet structure with optical waveguides has been fabricated using a commercial Fused Deposition Modeling (FDM) printer, enabling simultaneous actuation and deformation proprioception [22]. However, the optical path requirements of the waveguides generally impose constraints on the actuator geometry, favoring relatively planar structures and limiting the application of more complex designs, such as origami-inspired actuators [25]-[28]. Meanwhile, the  monolithic integration of force-proprioception and actuation using a single material remains to be explored.

In this work, we integrate an Asymmetric Origami Bending (AOB) chamber [29] with a Force-Proprioception Soft (FPS) sensor to propose a Monolithic Force-Proprioception Soft (MFPS) actuator. Using the conductive Thermoplastic Polyurethane (TPU) as the material, the proposed actuator integrates actuation and force sensing through a single material and a one-step manufacturing process. The main contributions are as follows:

\begin{enumerate}
\renewcommand{\labelenumi}{\arabic{enumi}.}
\item We propose the MFPS actuator design method that integrates force-proprioception and actuation using a single material and a one-step manufacturing process. This approach ensures consistent material properties across the continous actuator and sensor structures.

\item The sensing performance are investigated through structure design and parametric analysis, providing a readily detectable sensing signal. With printing parameters studied, the MFPS actuator is fabricated by FDM 3D-printing of conductive TPU. The actuation and sensing characteristics are experimentally characterized.

\item A two-finger force-proprioception gripper is developed. The gripper is used for grasping daily objects and sensing gripping force, demonstrating the effectiveness of the proposed MFPS actuator.
\end{enumerate}

In the following sections, Section II introduces the concept of the MFPS actuator; Section III introduces the structure of the AOB chamber and the design of FPS sensor; Section IV presents the detailed fabrication process of the MFPS actuator; Section V presents the actuator performance characterizations; Section VI demonstrates the practical applications, followed by the conclusions.

\section{MFPS Actautor Concept}
\begin{figure*}[htbp]
  \centering
\includegraphics[width=\textwidth]{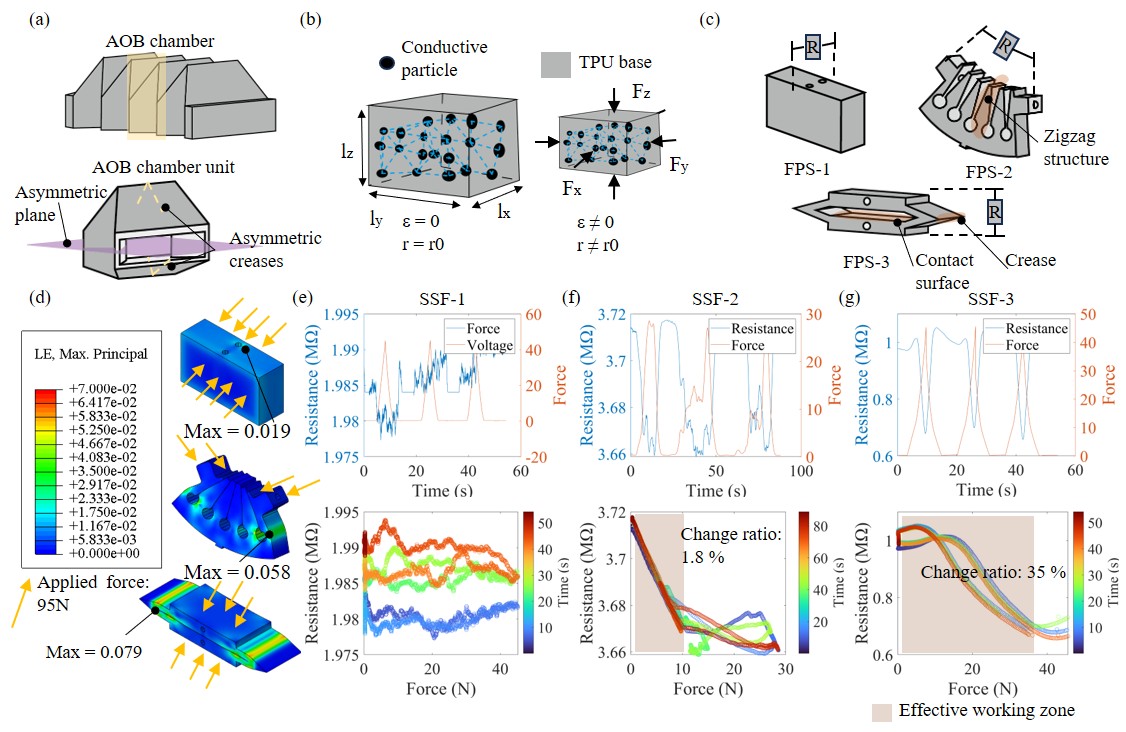}
  \caption{Design of the MFPS actuator. (a) The structure of the AOB chamber. (b) Sensing principle of the conductive TPU. (c) Structure design of the FPS sensors. (d) Simulation results of the FPS sensors. (e) Sensing performance of the FPS sensors.}
\end{figure*}
Human fingers serve as an important organ for human to interact with the environment, enabling a wide range of functions, including touching, sensing picking, grasping, etc. The finger inherently integrates actuation and proprioception within a compact biological structure, as illustrated in Fig. 1(a). The muscles and tendons underneath the skin generate and transmit mechanical forces to facilitate joint bending, while the skin covering the finger provides both physical interaction with the environment and mechanical perception. The skin consists of the epidermis and deeper tissue layers including tactile corpuscles which detect external mechanical stimuli and transduce local pressure into neural signals. This natural integration enables the finger to perceive the surrounding environment while simultaneously performing mechanical movements.

Inspired by this functional integration of actuation and proprioception in biological fingers, we propose the MFPS Actuator, as illustrated in Fig. 1(a). The MFPS actuator integrates an actuation module and a force-proprioception module within a continuous single soft structure. As illustrated in Fig. 1(b), the actuation module, which is realized by the AOB chamber, adopts an origami-inspired structure and generates the bending movement under the air pressure. As illustrated in Fig. 1(c), the force-proprioception module, realized by the FPS sensor, converts external force into an electrical signal through strain-induced resistance changes, thereby enabling force proprioception.

Overall, the proposed MFPS actuator concept is inspired by the integration of actuation and proprioception functions in biological fingers, aiming to achieve single-material, single-step fabrication of a soft actuator with integrated actuation and proprioception. By monolithically fabricating the actuation and sensing structures from the same material, interfacial stress concentrations caused by the bonding of different components and materials can be avoided.

\section{MFPS Actuator Design}
The MFPS actuator consists of the AOB chamber and the FPS sensor. In this section, the structures and working principles of the both are introduced.
\subsection{AOB Chamber}

\begin{figure*}[!t]
    \centering
    \includegraphics[width=\textwidth]{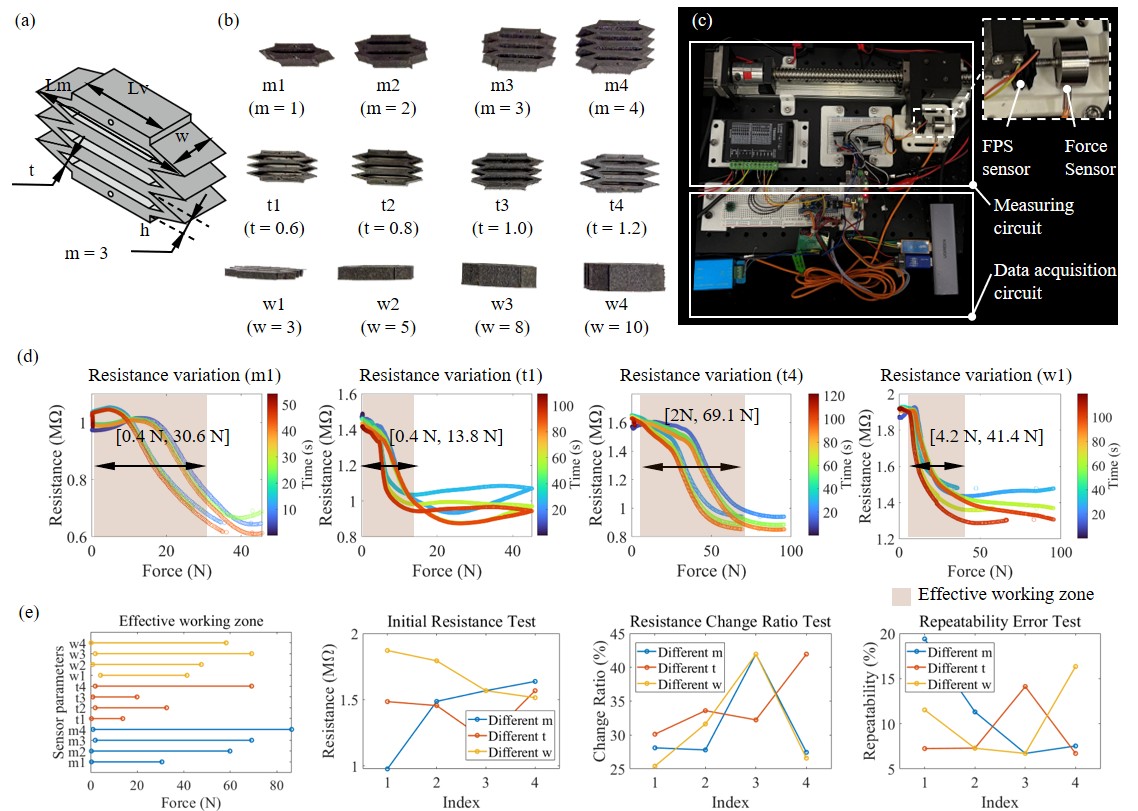}
    \caption{Parametric analysis of the FPS sensor. (a) Geometric description of the FPS sensor. (b) FPS sensor prototypes with different geometric parameters. (c) Experimental setup. (d) Force-resistance responses of four representative FPS sensor samples. (e) Sensing performance under varying geometric parameters, where the horizontal axis index corresponds to the parameter subscript (e.g., index 2 for parameter $t$ represents $t_2$, i.e., $t = 0.8\text{ mm}$).}
    \label{Different parameter}
\end{figure*}
The AOB chamber is evolved from the Yoshimura origami pattern. Based on the periodic creases of the Yoshimura origami pattern, an asymmetric geometric design is introduced to induce bending deformation during pneumatic inflation [29]. The AOB chamber consists of multiple AOB units arranged continuously, as shown in Fig. 2(a). Each unit incorporates asymmetric creases distributed at the opposite sides of the asymmetric plane. When air pressure is applied, the crease regions undergo corresponding unfolding, resulting in different degrees of deformation on the two sides of the structure and consequently inducing bending. By designing the asymmetric ratio and continuous arrangement of the units, the overall motion of the AOB chamber can be regulated. The specific AOB design parameters adopted in this work are identical as presented in [29].

\subsection{Sensing Principle}
The FPS sensor converts external mechanical deformation into a measurable signal, thereby enabling the proprioception of external forces. Its working principle is well suited to the material properties of conductive TPU. When the conductive TPU is subjected to an external mechanical force, the spatial distance between the conductive particles changes, thereby altering its electrical conductivity, as shown in Fig. 2(b).

For an undeformed conductive TPU unit with initial dimensions \(l_x\), \(l_y\), and \(l_z\), its initial resistivity \(r_0\) can be assumed to be 
\begin{equation}
r_0 = \frac{k}{\rho_0},
\label{eq1}
\end{equation}
where \(k\) represents the coefficient $\rho_0$ relationship between the conductive particle density and resistivity.

According to the generalized Hooke's law, under the assumption of linear elasticity, when the conductive unit is subjected to external forces $F_x$, $F_y$, and $F_z$, the normal strains in the three directions, $\varepsilon_x$, $\varepsilon_y$, and $\varepsilon_z$, can be expressed as
\begin{equation}
\begin{bmatrix}
\varepsilon_x \\
\varepsilon_y \\
\varepsilon_z
\end{bmatrix}
=
\frac{1}{E}
\mathbf{C}
\boldsymbol{\sigma},
\label{eq2}
\end{equation}
where $E$ is the Young's modulus, $\mathbf{C}$ denotes the compliance matrix. $\boldsymbol{\sigma}$ denotes the stress vector, given by
\begin{equation}
\mathbf{C}
=
\frac{1}{E}
\begin{bmatrix}
1 & -\nu & -\nu \\
-\nu & 1 & -\nu \\
-\nu & -\nu & 1
\end{bmatrix},
\qquad
\boldsymbol{\sigma}
=
\begin{bmatrix}
\dfrac{F_x}{l_y l_z} \\[10pt]
\dfrac{F_y}{l_x l_z} \\[10pt]
\dfrac{F_z}{l_x l_y}
\end{bmatrix},
\label{eq3}
\end{equation}
where $\nu$ is the Poisson's ratio of the conductive TPU. The density of the conductive particle under deformation can be expressed as 
\begin{equation}
\rho
=
\frac{\rho_0}
{
\det
\left[
\mathbf{I}
+
\operatorname{diag}
\left(
\mathbf{C}\boldsymbol{\sigma}
\right)
\right]
}.
\label{eq4}
\end{equation}
where I denotes the identical matrix $3\times 3$. Consequently, the deformed resistivity can be expressed as
\begin{equation}
r
=
r_0
\det
\left[
\mathbf{I}
+
\operatorname{diag}
\left(
\mathbf{C}\boldsymbol{\sigma}
\right)
\right].
\label{eq5}
\end{equation}
\subsection{FPS Sensor Structure}
From Eqns. \eqref{eq1}, \eqref{eq2}, and \eqref{eq4}, the change in resistivity induced by external force is related to the strain. Accordingly, three FPS sensors were designed to amplify the strain generated under compression, as shown in Fig. 2(c). FPS-1 is a cuboid structure, for which the compressive strain is primarily determined by the intrinsic material properties. FPS-2 adopts a zigzag structure. Under compression, deformation is concentrated at the roots of the zigzag features, while the adjacent surfaces progressively come into contact, thereby enhancing the compressive strain. FPS-3 incorporates creases into the square contact surface, concentrating strain around the creases while promoting contact between the upper and lower surfaces under compression.

The simulations were performed using Abaqus 2020. A general static step was defined, and the models were meshed using hexahedral elements (C3D8R). The external load of 95 N was applied to the corresponding surface, as shown in Fig. 2(d). The maximum strain within the FPS-1 structure is 0.019, while that within FPS-2 reaches 0.058, with the strain concentrated at the roots of the zigzag features. For FPS-3, the maximum strain within the structure occurs around the creases and reaches 0.079.

Through regulating the geometric parameters, the material consumptions of all three sensors were controlled to 1 g. The resistance of FPS sensors were tested (the experimental setup will be introduced in Subsection D). The results are shown in Fig. 2(e). To better characterize the sensor performance, we defined the effective working range, which is the force interval corresponding to 2\%-95\% of the total output voltage variation. For the FPS-1, the resistance is almost insensitive to the applied load, which can be attributed to the limited strain generated by the solid cuboid structure. FPS-2 exhibits a pronounced response to the applied load, particularly in the range of 0-10 N, where a relatively strong linear response is observed. However, its resistance changes by approximately 1.8\%, making it more susceptible to noise interference. Among the three designs, FPS-3 exhibits the best sensing performance, achieving a resistance change ratio of 35\% and a larger effective working range than the other two designs. 

In summary, the simulation and experimental results preliminarily validate that the strain developed in the structure has an influence on the resistance response. Owing to its performance, FPS-3 was adopted as the sensing structure of the FPS sensor.

\subsection{Parametric Analysis}
\begin{figure*}[!t]
    \centering
    \includegraphics[width=\textwidth]{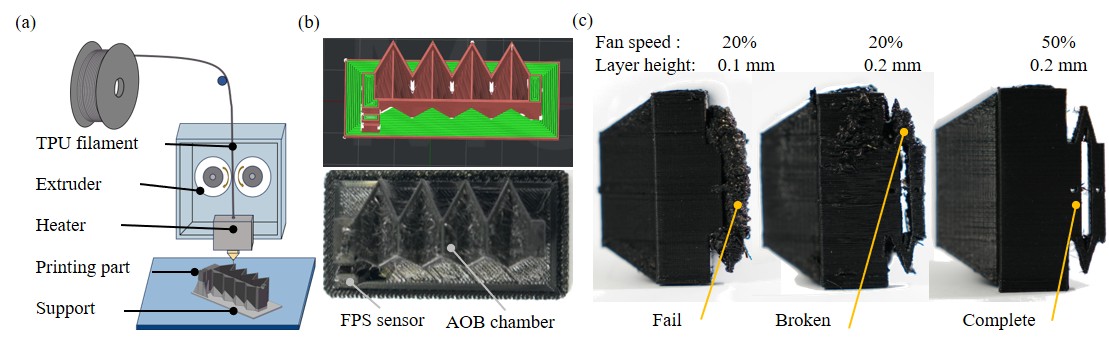}
    \caption{Fabrication of the MFPS actuator. (a) FDM 3D printing process. (b) Comparison of the sliced model and printed prototype.(c) Fabrication results under varying fan speeds and layer heights.}
    \label{Fabrication}
\end{figure*}
The FPS sensor can be described by six geometric parameters, as shown in Fig. 3(a). Here, \(L_v\) and \(L_m\) denote the distances between the two valley creases and between the two mountain creases located on opposite sides of the FPS sensor, respectively. \(t\) denotes the thickness of the crease, \(h\) denotes the height of a single FPS unit, \(w\) denotes the width of the FPS sensor and \(m\) denotes the number of serial units. Considering the structural parameters of the AOB chamber, \(L_m\), \(h\), and \(L_v\) are first determined to ensure structural compactness. The effects of \(m\), \(t\), and \(w\) on the sensing performance of the FPS sensor are then investigated.

Using FDM 3D-printing, a series of FPS sensors with distinct geometric configurations was fabricated by varying $m$ from 1 to 4 ($m_1\text{--}m_4$), $t$ from $0.6$ to $1.2\text{ mm}$ ($t_1\text{--}t_4$), and $w$ from $3$ to $10\text{ mm}$ ($w_1\text{--}w_4$), as shown in Fig. 3(b). The sensing performance of these prototypes were evaluated using the experimental setup presented in Fig. 3(c). In the measuring circuit, a ball-screw-driven linear stage was used to compress the FPS sensor, while a force sensor was employed to measure the actual applied force. The resistance variation of the FPS sensor was obtained through a voltage-divider circuit. The data acquisition circuit was used to obtain the voltage and force data. During the experiment, the linear stage advanced at a constant speed of 1 mm/s to compress the FPS sensor. When the measured force reached the preset value, the stage retracted, and the cycle was repeated three times with a 5 s interval between consecutive cycles, as demonstrated in Supplementary Movie S2.

Fig. 3(d) presents the force-resistance responses of four representative sensors among the fabricated samples. The FPS sensors exhibit hysteresis, primarily due to the inherent viscoelasticity of TPU. The hysteresis can be suppressed by appropriately adjusting the geometric parameters, as demonstrated by the resistance response of the sensor with \(w=3\). The effective working range is also affected by the geometric parameters. For example, the upper limit of the effective working zone reaches 69.1 N for \(t=1.2\).

Fig. 3(e) summarizes the characteristics obtained from the experiments. The results show that the effective working zone increases approximately linearly with \(m\). At \(m=4\), the effective working zone reaches \([1.0 N,86.3 N]\). The parameter \(t\) and \(m\) have a large effect on the initial resistance, with the minimum initial resistance obtained at \(m=1\). The largest change ratio obtained at \(t=1.2\), reaching 43\%. Among the investigated parameters, the minimum repeatability error is approximately 7\%. Considering the sensing performance, the final geometric parameters of the FPS sensor were determined, as listed in Table I.

\begin{table}[t]
\centering
\caption{Geometric parameters of the FPS sensor.}
\label{tab:fps_parameters}
\begin{tabular}{cccccc}
\toprule
\(L_v\) (mm) & \(L_m\) (mm) & \(t\) (mm) & \(h\) (mm) & \(w\) (mm) & \(m\) \\
\midrule
15 & 29 & 1.2 & 6.4 & 5 & 1 \\
\bottomrule
\end{tabular}
\end{table}

\section{Experimental Characterization}
In this section, we investigate the actuation and proprioception performance of the MFPS actuator.

\subsection{Fabrication}
\begin{figure*}[htbp]
  \centering
  \includegraphics[width=\textwidth]{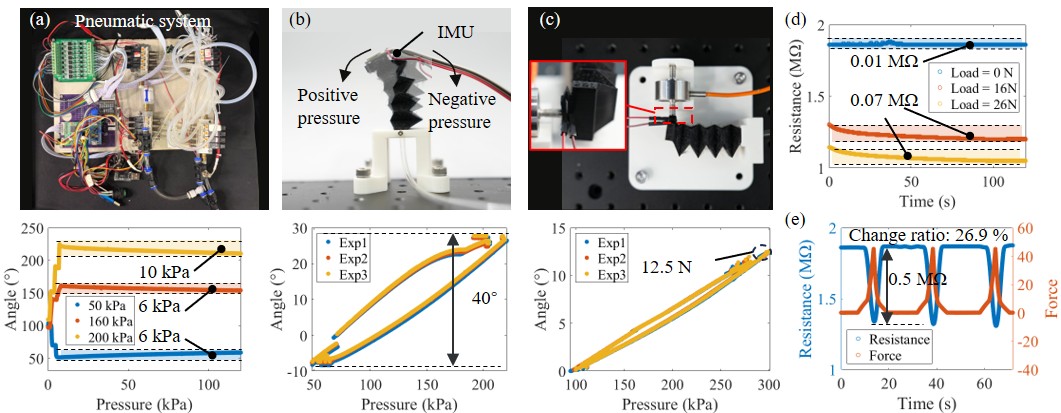}
  \caption{Experimental characterization of the MFPS actuator. Experimental setup and results of the (a) airtightness characterization, (b) bending displacement characterization and (c) output force characterization. (d) Zero drift of the MFPS actuator. (e) Force-proprioception characterization of the MFPS actuator. }
\end{figure*}
All samples in this study were fabricated using commercial conductive TPU filament with an FDM 3D printer (Raise3D Pro2 Plus, Raise3D Technologies, Inc.), as shown in Fig. 4(a). To achieve monolithic single-material fabrication, a dedicated support structure was designed to ensure successful printing. The sliced model and fabricated sample are compared in Fig. 4(b). The MFPS actuator can be fabricated entirely by FDM printing and used after support removal, while the sensor only requires wire connection for signal acquisition, as demonstrated in Supplementary Movie S1.

During fabrication, the fan speed and layer height affect the forming quality of the MFPS actuator. Although a lower fan speed and smaller layer height are beneficial for improving the airtightness of the actuator, an excessively low fan speed prevents the extruded material from solidifying in time, while an excessively small layer height increases the friction between the nozzle and the printed material, causing the relatively weak sensor region to move with the nozzle during printing. These factors can lead to FPS sensor printing failure and loss of sensing functionality, as shown in Fig. 4(c).

\subsection{Airtightness verification}
As shown in Fig. 5(a), a pneumatic system was utilized for the airtightness test. The solenoid valve was controlled by an STM32 microcontroller to regulate the actuator to the desired pressure. To evaluate the airtightness of the MFPS actuator, the internal pressure was measured at 220, 160, and 50 kPa, with atmospheric pressure of 101 kPa as the reference. Over a 120 s test period, the pressure decreased by 10, 6, and 6 kPa, respectively. These results demonstrate that the MFPS actuator maintains sufficient airtightness.
\subsection{Bending Displacement}
As shown in Fig. 5(b), the bending displacement of the actuator was measured using an IMU (Inertial Measurement Unit). The actuator was first slowly depressurized to 50 kPa and held for 5 s, followed by pressurization to 220 kPa and another 5 s hold before returning to atmospheric pressure. This cycle was repeated three times, while the bending angle was recorded in real time by the IMU. The actuator exhibited a bending angle of approximately $-10^\circ$ at 50 kPa and $30^\circ$ at 220 kPa. The actuation capability of the MFPS actuator was demonstrated in Supplementary Movie S3. These results demonstrate the appropriate active bending capability of the MFPS actuator.

\subsection{Output Force}
As shown in Fig. 5(c), the output force of the MFPS actuator was characterized by gradually increasing the pressure to 300 kPa using the pneumatic system. The test was repeated three times, and the output force was recorded in real time using a force sensor. The results show that the output force exhibits an approximately linear relationship with the applied pressure, reaching 12.5 N at 300 kPa.
\subsection{Proprioception Characterization}
The proprioception performance of the MFPS actuator was characterized using the experimental setup shown in Fig. 3(c). First, the zero drift of the MFPS actuator was evaluated under different loading conditions, as shown in Fig. 5(d). Under no-load conditions, the zero drift was 0.01 M$\Omega$, corresponding to 0.05\%. At applied loads of 16 and 26 N, the zero drift increased to 0.07 M$\Omega$, corresponding to 5.0\% and 5.6\%, respectively. These results indicate that the sensing response is affected by the applied external load. 

The resistance response of the MFPS actuator was characterized under repeated loading up to 40 N, as shown in Fig. 5(e). The MFPS actuator exhibited appropriate repeatability under cyclic loading from 0 to 45 N, demonstrating a stable force-sensing response. The resistance change achieves 0.5 M$\Omega$ with change ratio of 26.9\%.

\section{Two-finger Force-proprioception Gripper}
\begin{figure*}[htbp]
  \centering
  \includegraphics[width=\textwidth]{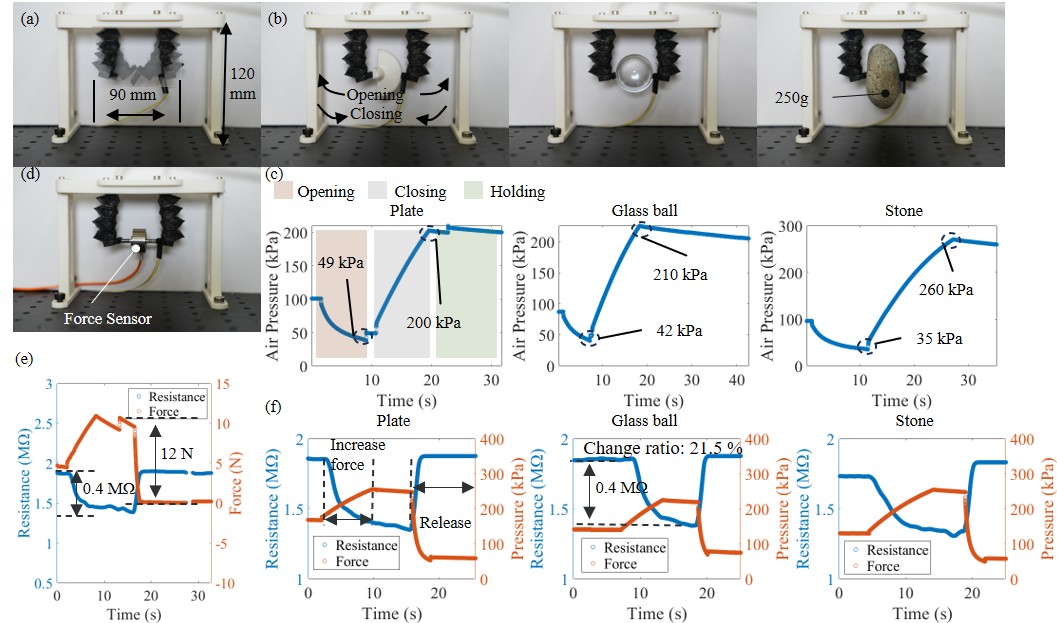}
  \caption{Soft gripper application. (a) two-finger force-proprioception gripper with a gripping range of 0--90 mm. (b) Gripping demonstrations and (c) air pressure variation during the gripping process. (d) Experimental setup for measuring the gripping force of the gripper and (e) corresponding gripping force results.(f) Relationship between the resistance and air pressure signals during increasing grasping force.}
\end{figure*}
In this section, a two-finger force-proprioception gripper is developed using the MFPS actuators to demonstrate its potential for gripping and force proprioception.
\subsection{Grasping Demonstration }
As shown in Fig. 6(a), the two-finger force-proprioception gripper consists of two MFPS actuators and has a gripping range of 0-90 mm. The gripper successfully grasped a plate, glass ball, and a stone, demonstrating its ability to grasp objects with different shapes and weights, as shown in Fig. 6(b) and Supplementary Movie S4. The pressure sequence during grasping is shown in Fig. 6(c). Negative pressure was first applied to open the gripper, followed by positive pressure to close the fingers and hold the grasp for 5 s. These results demonstrate the effective actuation capability of the MFPS-based two-finger gripper.
\subsection{Proprioception demonstration}
The gripping force was measured using a force sensor while the resistance response of the integrated FPS sensors was recorded. As shown in Fig. 6(d) and Supplementary Movie S5, positive pressure was gradually applied to the gripper, which was then held at the target pressure for 10 s before returning to atmospheric pressure. At an applied pressure of 220 kPa, the gripper generated a gripping force of approximately 12 N, accompanied by a resistance change of 0.4 M$\Omega$, as shown in Fig. 6(e). The force-proprioception capability was also evaluated during the grasping of a plate, a glass ball, and a stone. For each object, the pressure was increased to approximately 250 kPa to increase the grasping force and maintained for approximately 5 s, while the resistance response was recorded. As shown in Fig. 6(f), the results show a clear negative correlation between the resistance and air pressure during grasping. The resistance change achieves 0.4 M$\Omega$, with change ratio of 21.5\%. These results demonstrate the effective force-proprioception capability of the two-finger MFPS gripper.

\section{Discussion and Conclusion}
Inspired by the functional integration of actuation and sensing in human finger, we propose the MFPS design method that integrates actuation and force-proprioception using a single material under one-step fabrication. An MFPS actuator consisting of an AOB chamber and an FPS sensor was designed, fabricated, and characterized. A two-finger force-proprioception gripper was developed for validation.

The FPS sensor was structurally designed and evaluated, with FPS-3 selected based on simulation and experimental results. Parametric analysis demonstrated that the resistance change rate could be effectively regulated by the sensor geometry, reaching a maximum of 43\%. The MFPS actuator was successfully fabricated by FDM and achieved a bending angle of $40^\circ$, an output force of $12.5$~N, a resistance zero-point drift of $0.05\%$, and a resistance change of $26.9\%$ as the applied external force increased from 0 to 45 N. Furthermore, a two-finger force-proprioception gripper was developed based on the MFPS actuator, achieving an output force of 12 N and a resistance change rate of 21.5\% during the grasping of daily objects.

In summary, the proposed MFPS design method effectively integrates the actuation and force-proprioception functions of a pneumatic soft actuator using a single material. The MFPS actuator demonstrates appropiate actuation performance and promising force-proprioception capability. However, the FPS sensor exhibits noticeable hysteresis, and further investigation is required to reduce the hysteresis through geometric design. In addition, the influence of the AOB chamber on the sensing performance after integrating the FPS sensor into the MFPS actuator will be investigated in future work.

\vspace{12pt}
\end{document}